\documentclass[10pt,twocolumn,letterpaper]{article}

\usepackage{iccv}              
\usepackage{ulem}

\usepackage{xcolor}
\definecolor{yellow_color}{RGB}{255,200,0}
\definecolor{blue_color}{RGB}{0,0,255}
\definecolor{red_color}{RGB}{255,0,0}

\usepackage{multirow}
\usepackage{pifont}
\usepackage{cuted}
\usepackage{xcolor}

\newcommand{\cmark}{\color{green} \ding{51}} 
\newcommand{\xmark}{\color{red} \ding{55}} 

\definecolor{iccvblue}{rgb}{0.21,0.49,0.74}
\usepackage[pagebackref,breaklinks,colorlinks,allcolors=iccvblue]{hyperref}
\usepackage{tikz}

\def\paperID{5060} 
\def\confName{ICCV}
\def\confYear{2025}

\title{DexH2R: A Benchmark for Dynamic Dexterous Grasping \\in Human-to-Robot Handover}

\author{
\textbf{
Youzhuo Wang$^{1,}$\thanks{Equal contribution. $\dagger$ Corresponding author.}  , 
Jiayi Ye$^{1,}\footnote[1]{}$  , 
Chuyang Xiao$^1$, 
Yiming Zhong$^1$, 
Heng Tao$^1$, 
Hang Yu$^1$, 
}\\
\textbf{
Yumeng Liu$^{1,2}$, 
Jingyi Yu$^1$, 
Yuexin Ma$^{1\dag}$}\\
$^1$ ShanghaiTech University
$^2$The University of Hong Kong\\
{\tt\small \{wangyzh2023,yejy2024,mayuexin\}@shanghaitech.edu.cn}}

\begin{document}
\maketitle
\begin{strip}
    \centering
    \vspace{-5.2em}
    \centering
    \includegraphics[width=\linewidth]{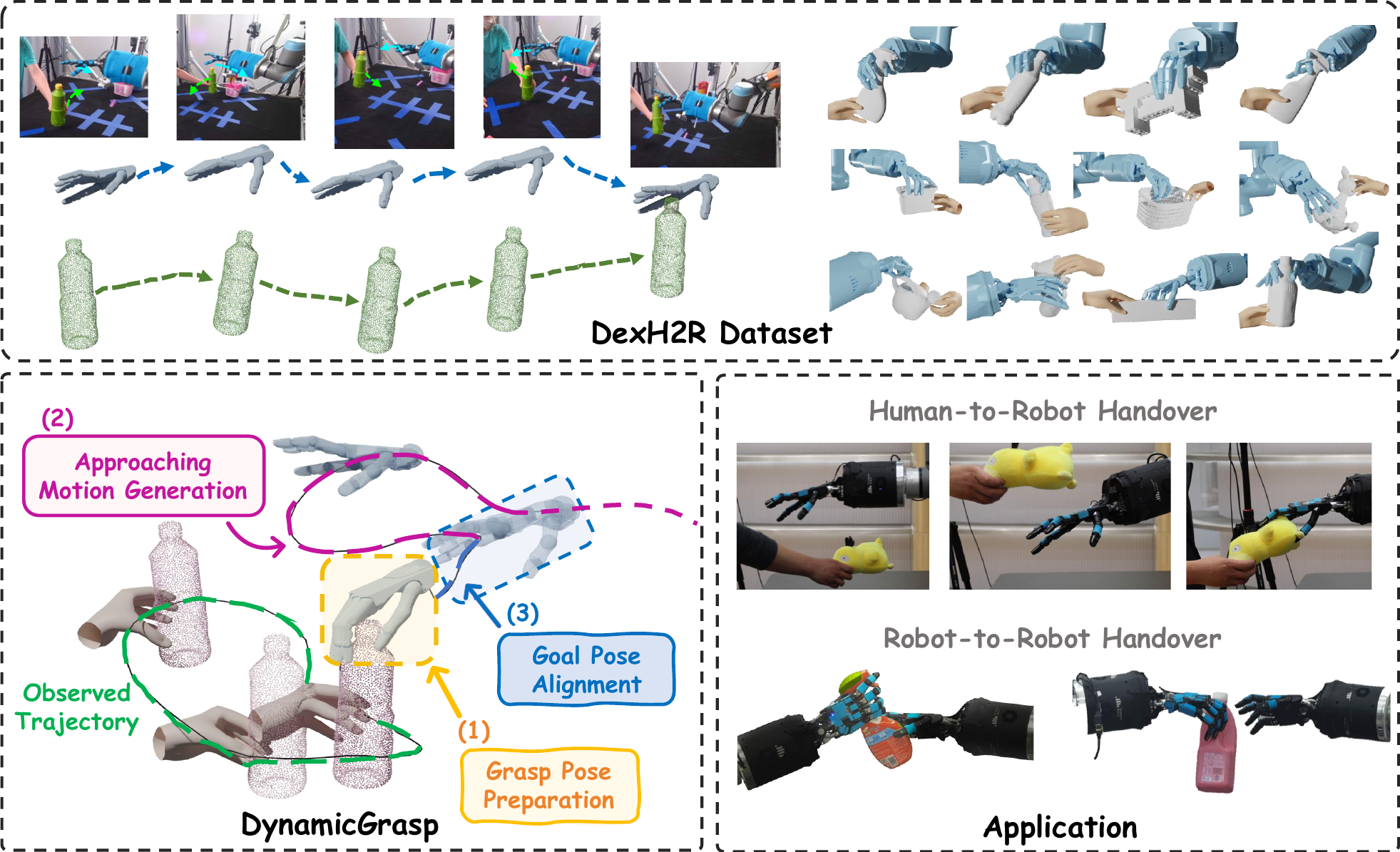}
    \vspace{-1em}
    \captionof{figure}{We introduce \textbf{DexH2R}, the first real human-to-robot handover dataset, and establish an effective and practical solution \textbf{DynamicGrasp} along with a comprehensive benchmark for human-to-robot handover. This benchmark can benefit broad real-world applications and can be extended to robot-to-robot handover tasks.}
    \vspace{-1em}
    \label{fig:teaser}
\end{strip}

\begin{abstract}
Handover between a human and a dexterous robotic hand is a fundamental yet challenging task in human-robot collaboration. It requires handling dynamic environments and a wide variety of objects and demands robust and adaptive grasping strategies. However, progress in developing effective dynamic dexterous grasping methods is limited by the absence of high-quality, real-world human-to-robot handover datasets. Existing datasets primarily focus on grasping static objects or rely on synthesized handover motions, which differ significantly from real-world robot motion patterns, creating a substantial gap in applicability.
In this paper, we introduce DexH2R, a comprehensive real-world dataset for human-to-robot handovers, built on a dexterous robotic hand. Our dataset captures a diverse range of interactive objects, dynamic motion patterns, rich visual sensor data, and detailed annotations.  
Additionally, to ensure natural and human-like dexterous motions, we utilize teleoperation for data collection, enabling the robot's movements to align with human behaviors and habits, which is a crucial characteristic for intelligent humanoid robots.
Furthermore, we propose an effective solution, DynamicGrasp, for human-to-robot handover and evaluate various state-of-the-art approaches, including auto-regressive models and diffusion policy methods, providing a thorough comparison and analysis.  
We believe our benchmark will drive advancements in human-to-robot handover research by offering a high-quality dataset, effective solutions, and comprehensive evaluation metrics. \textbf{Project is at \href{https://dexh2r.github.io/}{dexh2r.github.io/}.}
\end{abstract}

\section{Introduction}

Enabling a robot to receive objects handed over by a human naturally is a fundamental operation in human-robot interaction and collaboration. This capability has far-reaching applications across various domains, including household services, healthcare, industrial automation, etc. Compared to two-finger grippers, five-finger dexterous robotic hands more closely mimic the structure and functionality of the human hand, allowing for more sophisticated, human-like operations and behaviors. As a result, they have attracted more and more attention in intelligent robotics research. However, their high degrees of freedom, along with complex geometric and physical constraints, pose significant challenges for human-to-robot handover.

Achieving natural and seamless human-to-robot handover requires addressing several key factors. \textit{Safety} is paramount, as the robot must anticipate and respond to human movements to prevent risks, integrating collision avoidance, especially in close proximity~\cite{sisbot2007spatial,eguiluz2017reliable}. \textit{Human-like behavior} enhances intuitive collaboration by mimicking human motion patterns, reducing cognitive load, and fostering trust~\cite{cakmak2011human}. \textit{Practicability} ensures real-world applicability, requiring learning from high-quality data while maintaining strong generalization capabilities~\cite{liu2023target}.  
To overcome these challenges and enable effective dynamic dexterous grasping, high-quality real-world data, and practical solutions are essential.

Existing datasets struggle to capture realistic human-to-robot interactions. Most dexterous grasping datasets~\cite{wang2023dexgraspnet, liu2024realdex, hang2024dexfuncgrasp} focus on static object grasping, lacking the dynamic sequences crucial for handover motions. Some recent works attempt to synthesize handover data by augmenting static datasets. For example, HandoverSim~\cite{chao2022handoversim} integrates the motion-captured DexYCB dataset~\cite{chao2021dexycb} into simulators, while GenH2R~\cite{wang2024genh2r} and MobileH2R~\cite{wang2025mobileh2r} generate synthetic human handover motions using parametric 3D animation trajectories.  
However, due to the use of synthetic givers, receivers, and visual input, these approaches fail to accurately capture the real motion patterns of human-to-robot handovers, creating a gap between simulation and real-world applications. Additionally, they primarily focus on two-finger robotic systems, making them inadequate for five-finger dexterous hands. These limitations highlight the urgent need for high-quality datasets that authentically represent dynamic human-to-robot handovers.

To overcome these limitations and move beyond synthetic data, we introduce \textbf{DexH2R}, the first real-world human-to-robot handover dataset designed specifically for dexterous robotic hands. DexH2R is distinguished by three key characteristics: 1) \textbf{Rich Multi-Modal Perception Data}, 2) \textbf{Authentic Human-Object-Robot Interaction}, and 3) \textbf{Human-like Dynamic Grasping Behavior}.
These features are achieved through a comprehensive visual sensing system that captures high-quality multi-modal perception data with precise 3D annotations of objects, human hands, and robotic hands. Additionally, we ensure authentic human-object-robot interactions by employing the ShadowHand teleoperation system~\cite{ShadowRobot}, which enables the robotic hand to synchronize its movements with a human operator in real-time, facilitating natural and human-like grasping behavior.
Our dataset comprises 4282 handover trials across 39 participants and 56 objects spanning diverse shapes, offering a valuable resource for training and evaluating human-to-robot handover methods.

Based on our dataset, we propose an effective and applicable solution, \textbf{DynamicGrasp}, for the human-to-robot handover task. Reflecting on how humans receive objects, the process follows three key stages: First, the system anticipates an optimal grasping position and posture. Next, it continuously refines its expectations and adjusts the approaching motion based on the giver’s movements. Finally, as the robot nears the object, it shifts its focus to achieving a successful and secure handover pose. 
Inspired by human behaviors, our method consists of three key stages: (1) \textbf{Grasp Pose Preparation}, (2) \textbf{Approaching Motion Generation}, and (3) \textbf{Goal Pose Alignment}.
First, to ensure the generalization capability and physical plausibility of grasping poses, we pre-train a grasp pose generation model using a large-scale simulated dataset combined with our real-world data, incorporating a physically constrained filtering process.  
Second, we introduce two technical paradigms for approaching motion generation: autoregressive approaches~\cite{liu2024realdex} and diffusion-policy approaches~\cite{chi2024diffusionpolicy,ze20243d}, providing a comprehensive comparison and analysis of their performance.  
Third, we adopt a simple but effective linear interpolation strategy to align with the final safe goal pose, ensuring a successful receiving motion, which enhances applicability and reliability in real-world experiments. Finally, by leveraging both our human-like dynamic grasping data and solution, we equip the dexterous hand with human-like intelligence.

Additionally, we design specific evaluation metrics to assess our solution across key aspects such as safety, precision, and reliability. Extensive experiments further validate our approach, offering valuable insights into its strengths and limitations in human-to-robot handover scenarios. We sincerely hope that our valuable dataset, solution, evaluation, and discussion will establish a comprehensive and meaningful benchmark for the human-to-robot handover task, contributing positively to the advancement of the human-robot interaction field.

\section{Related Work}

\subsection{Dataset for Handover}
Recent studies have made significant strides in establishing standardized settings for handovers. To support these efforts, various datasets have been introduced to capture the full handover process. For example, H2O~\cite{ye2021h2o} and HOH~\cite{wiederhold2024hoh} are human-to-human handover datasets that include detailed hand and object annotations.
In contrast, some datasets focus on specific aspects of the handover process rather than capturing it in its entirety. For instance, datasets such as ContactPose~\cite{brahmbhatt2020contactpose} and DexYCB~\cite{chao2021dexycb} emphasize the giving motion of human hands, while RealDex~\cite{liu2024realdex} focuses on robotic hands grasping static objects in real-world settings. These datasets are valuable for modeling the giver's behavior in human-to-robot handover scenarios, especially in simulation environments like HandoverSim~\cite{chao2022handoversim}, which has become a widely adopted virtual testbed in recent human-to-robot handover research~\cite{christen2023learning, christen2024synh2r}.
To further benchmark human-to-robot handover tasks, more synthetic data in simulators, such as GenH2R~\cite{wang2024genh2r} and MobileH2R~\cite{wang2025mobileh2r}, have also been proposed. However, these datasets are constrained by their reliance on parallel grippers as receivers, making them unsuitable for studying five-finger dexterous hands. Moreover, their simulated environments and data lack real visual data and handover motion patterns, causing the gap between simulations and real-world conditions.
In this work, we introduce DexH2R, the first dataset to capture human-to-robot handovers involving dexterous robotic hands in real-world scenarios with rich visual and annotation data.

\subsection{Dynamic Grasping}
Grasping a moving object is particularly challenging, as the robot must continuously predict and adapt to the object's future locations and poses. Traditional motion planning methods, ~\cite{zucker2013chomp,park2012itomp,schulman2014motion} have been widely used to generate smooth and optimized trajectories for robotic manipulation. However, these approaches often struggle with real-time adaptability in highly dynamic environments.
Recently, some studies have explored dynamic grasping with grippers for objects following simple and predictable trajectories~\cite{akinola2021dynamic,burgess2022dgbench,jia2024dynamic}. Other studies have explored more complex tasks, including dexterous hand coordination for throwing and catching, though primarily with simple geometric objects like balls, cubes, or rods~\cite{huang2023dynamic}. Another study proposes a reactive control strategy for dynamic dexterous grasping, which tracks the environment and synthesizes grasps within an open-loop framework~\cite{duan2024reactive}. In this work, we propose a dataset and solution for dynamic dexterous grasping in particular for the human-to-robot handover scenarios.


\subsection{Dexterous Grasping}
Dexterous hands offer unmatched flexibility and adaptability over traditional grippers, enabling robots to perform complex tasks in human-centric environments~\cite{mattar2013survey,mandikal2022dexvip}. However, their intricate structure poses significant challenges for effective grasping. 
Traditional model-based control strategies~\cite{kumar2016optimal,nagabandi2020deep,andrews2013goal} rely on manually designed rules, making it difficult to handle objects of varying shapes. To overcome these limitations, recent studies have explored imitation learning, which requires large-scale, high-quality demonstrations~\cite{mandikal2022dexvip,qin2022dexmv,ye2023learning,chen2022dextransfer}. Alternatively, some works have focused on reinforcement learning (RL), enabling robots to acquire grasping skills without relying on reference grasps~\cite{mandikal2021learning,xu2023unidexgrasp,wan2023unidexgrasp++,zhang2024graspxl}. However, RL training often struggles to converge and suffers from limited generalization ability. 
To make the model learn more real human behaviors and real dexterous robotic hand actions, supervised learning is necessary before real-world deployment and application. There are two dominant technology paradigms, including autoregressive methods~\cite{liu2024realdex} and diffusion policies~\cite{chi2024diffusionpolicy,chi2024diffusionpolicy,graspanyting}. The former is easy to implement and performs stable.
The latter uses a generative model for visuomotor policy learning, which scales well to high-dimensional output spaces, enabling the prediction of action sequences rather than single steps. Our method evaluates diverse technologies inside our whole solution pipeline to solve human-like dexterous grasping.

\section{Dataset}
\label{sec:dataset}

\begin{table*}[!t]
\centering
\resizebox{\linewidth}{!}{
\begin{tabular}{@{}cccccccccccc@{}}
\toprule
\multirow{2}{*}{Dataset} & \multirow{2}{*}{Realistic} &
  \multicolumn{3}{c}{Handover} &
  \multicolumn{4}{c}{Real Visual Data} &
  \multirow{2}{*}{\#Object} &
  \multirow{2}{*}{\#Subject} &
  \multirow{2}{*}{\#Interaction} \\ 
\cmidrule(lr){3-5} \cmidrule(lr){6-9}
& & type & giver      & receiver   & rgb & depth & ego & \#view &     &    &      \\ 
\midrule
H2O~\cite{ye2021h2o}  &\cmark   & H2H  & MANO  & MANO  & \cmark & \xmark & \xmark  & 5  & 30  & 15 & 1200 \\
HOH~\cite{wiederhold2024hoh} &\cmark    & H2H  & MANO       & MANO       &\cmark & \cmark   & \xmark  & 8       & 136 & 40 & 2720 \\ 
\midrule
*DexYCB~\cite{chao2021dexycb} &\cmark &  H2X   & MANO & -  & \cmark & \cmark   & \xmark  & 8  & 20  & 10 & 1000 \\
*HandoverSim~\cite{chao2022handoversim} & \xmark & H2X & MANO & - & \xmark & \xmark   & \xmark  & - & 20  & 10 & - \\
*RealDex~\cite{liu2024realdex} &\cmark & R2X  &  ShadowHand & -  & \cmark & \cmark  &\xmark  & 4 & 52  & 1  & 2630 \\ 
\midrule
GenH2R~\cite{wang2024genh2r} &\xmark &H2R &MANO &  Gripper  & \xmark & \xmark   & \xmark  & -  & 20  & 10 & - \\
Ours   &\cmark &H2R &MANO &  ShadowHand &\cmark &\cmark   & \cmark & 18      & 56  & 39 & 4282 \\ \bottomrule
\end{tabular}
}
    \caption{Comparison with existing handover datasets. Entries marked with * denote datasets that can serve as givers to generate synthetic handover data or act as benchmarks. H2H denotes human-to-human, H2R denotes human-to-robot, H2X denotes human-to-any, and R2X denotes robot-to-any. \# indicates attribute count. - represents unavailable data. }
    \label{tab:dataset}
\end{table*}

Bridging the lack of real-world hand-to-robot handover data, we introduce \textbf{DexH2R}—a dataset of 4282 human-to-robot handover trials with 456K frames collected from 39 participants and 56 everyday objects. The dataset includes 79911 successful final grasp poses achieved through teleoperated control of a ShadowHand robotic system. Apart from high-fidelity motion data, DexH2R also provides multi-view RGB-D streams from fixed and egocentric perspectives, along with 3D annotations of the human hand and objects, all with temporal synchronization across sensory modalities.

\subsection{Hardware System Setup}
\label{sec:hardware_system}
\begin{figure}[t]
    \centering
    \includegraphics[width=\linewidth]{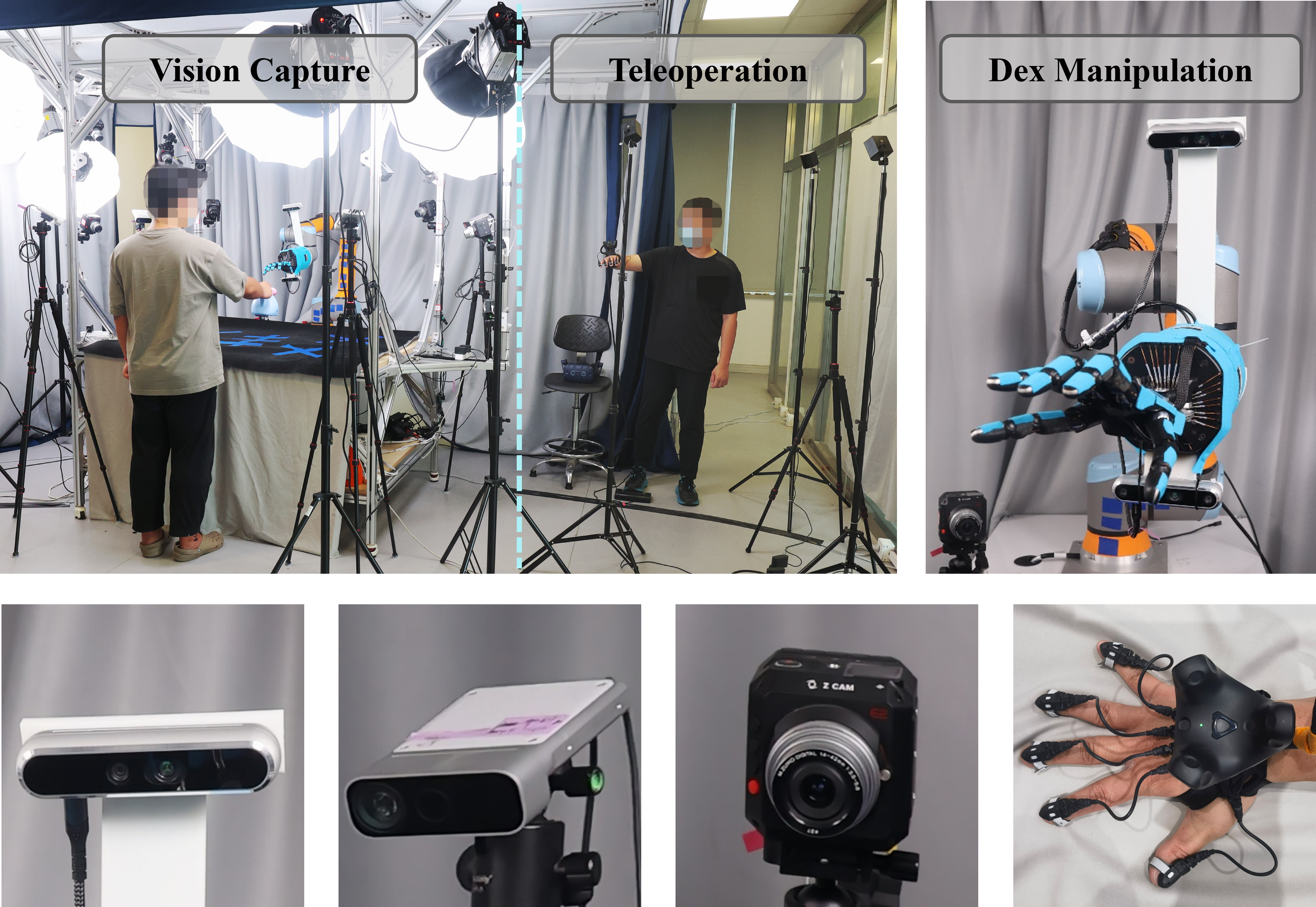}

    \caption{Overview of our hardware system. The upper part shows the system setup during dataset recording, while the lower part highlights the key components: RealSense D455, Azure Kinect, ZCAM E2, and the teleoperation glove.}
    \vspace{-1em}
    \label{fig:system_setup}
\end{figure}

Our data collection system comprises two key components: a vision capture system for environment tracking and a manipulation system utilizing teleoperation for controlling the ShadowHand's motions, as illustrated in~\cref{fig:system_setup}.

\noindent\textbf{Vision Capture System}. To capture detailed and diverse perspectives of the handover scene, we utilize a multi-view vision system comprising 12 cameras arranged in a quarter-sphere layout on the human giver's side of the workspace. This setup provides high-quality RGB streams and ensures 360-degree coverage of the handover interaction. 
In addition, 4 Azure Kinect cameras are placed at the corners of the workspace for the 3D reconstruction of the handover process. Beyond static viewpoints, we also incorporate ego-centric views by mounting two RealSense cameras on the wrist of the robotic arm using custom 3D-printed fixtures. One is positioned below the wrist level and the other is slightly elevated above the ShadowHand.

\noindent\textbf{Dexterous Manipulation System}. 
Our dexterous manipulation system pairs a UR10e robotic arm with a right ShadowHand, a tendon-driven robotic hand with 24 DoF. To achieve natural and realistic handover motions, we utilize a \textless 50ms-latency teleoperation system that transmits hand motions from a human operator, enabling the manipulation system to seamlessly interact with human givers.

\begin{figure}[hpt]
    \centering

    \includegraphics[width=\linewidth]{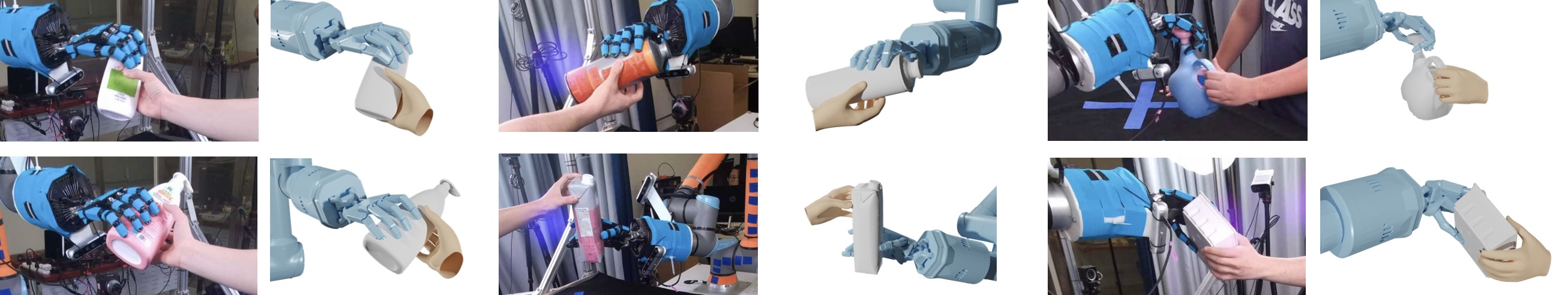}
    \caption{Examples of 3D annotation data in DexH2R.}
    \vspace{-1em}
    \label{fig:dataset}

\end{figure}

\subsection{Data Processing}

As~\cref{fig:dataset} shows, to facilitate different studies of dynamic grasping, we conduct precise annotation of object poses and human hand poses on raw visual data, providing valuable information for both object-centric and scene-level reasoning tasks.
We first apply SAM2~\cite{ravi2024sam} to segment object regions from RGB-D streams, we use the object region to do depth fusion so that we locate the object in the full scene and crop the object point cloud from it. 
With the located object region, we employed global registration~\cite{zhou2018open3d} to estimate the object's pose in the first frame. Then we inspect and correct the results manually. Subsequently, we used a frame-by-frame ICP method~\cite{classical_icp} to annotate the object poses on the processed clean point cloud. For hand poses, we detect 2D key points on the RGB images from ZCAM using MediaPipe~\cite{lugaresi2019mediapipe} and then estimate the MANO~\cite{MANO:SIGGRAPHASIA:2017} parameter of hand by off-the-shelf system.

\begin{figure*}[t]
    \centering
    \includegraphics[width=\linewidth]{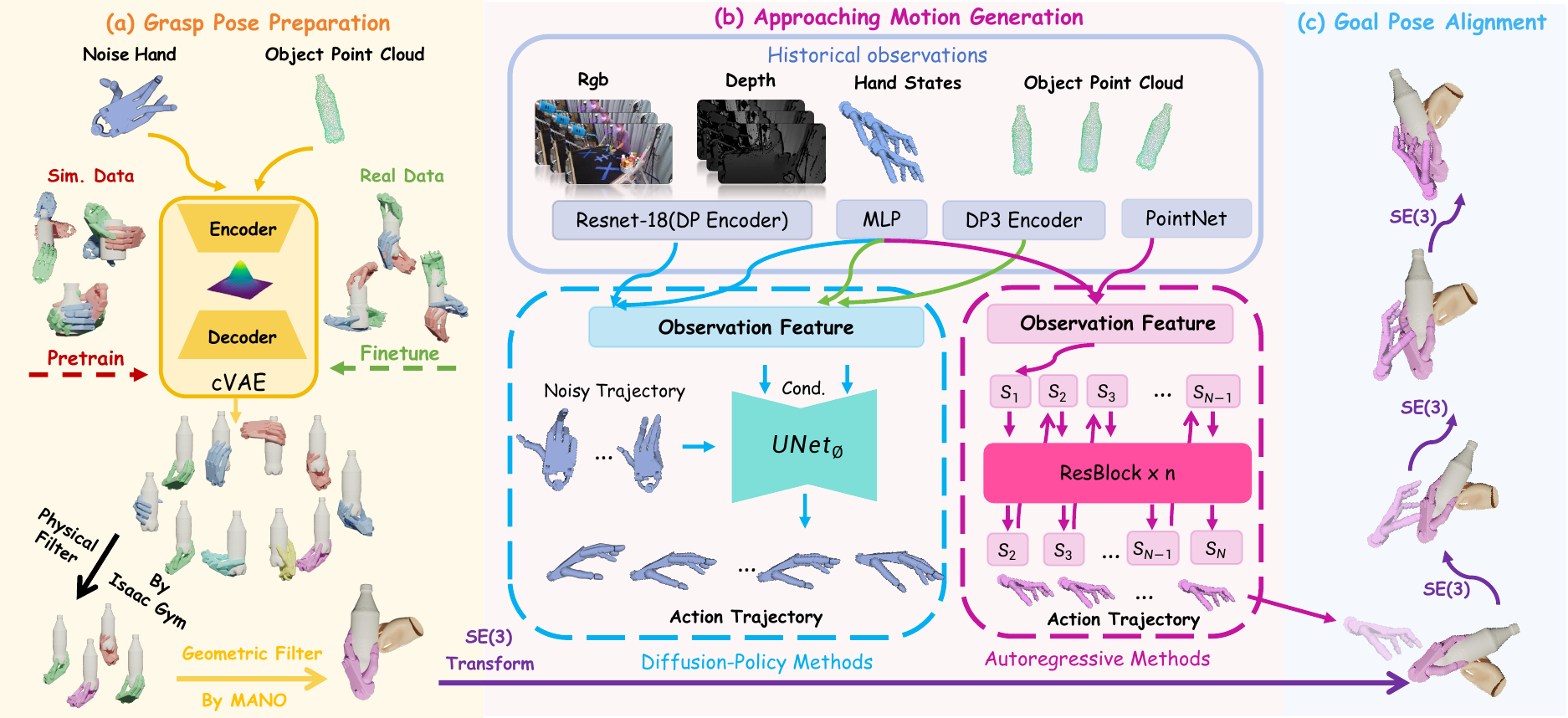} 
  
    \caption{Our dynamic grasping solution for human-to-robot handover consists of three stages: (a) grasping pose preparation, (b) approaching motion generation, and (c) goal pose alignment. In the first stage, we pre-train a grasping pose generation model on large-scale simulation data and fine-tune it with real-world grasping data. The generated pose candidates are refined using physical and geometric filters to guarantee a stable and practical grasping pose. For approaching motion generation, we explore both diffusion-policy and autoregressive methods to generate sequential poses. Once the hand reaches a predefined proximity to the object, the system transitions to the final stage. In the goal pose alignment stage, we employ a simple yet effective linear interpolation method to ensure precise and physically plausible grasping, maintaining accuracy and adherence to environmental constraints for a robust and reliable grasping outcome.}
    \label{fig:pipeline}
  
\end{figure*}

\subsection{Characteristics}

Compared to previous human-to-robot handover datasets in~\cref{tab:dataset}, DexH2R offers three key characteristics that enable more advanced research and have the potential to drive progress in this field.

\noindent\textbf{Rich Multi-Modal Perception Data} 
DexH2R dataset offers rich multimodal perception data with extensive multi-view coverage and detailed geometric representations. It includes depth information from six views and RGB images from 18 views, comprising 16 static external views and two ego-centric views. The fixed external cameras capture the entire handover process, documenting consistent human and robot motion within the scene, while the ego-centric cameras mounted on the ShadowHand provide close-up, detailed views of the target object during grasping. In addition, the dataset features high-resolution point clouds and accurately annotated meshes of human hands, a dexterous robotic hand, and objects, enabling fine-grained interaction analysis. By supporting diverse sensor configurations and multimodal observations, the dataset facilitates in-depth research on human-to-robot handovers.

\noindent\textbf{Authentic Human-Object-Robot Data.}
DexH2R offers realistic human handover behavior, authentic object states, and genuine robotic dynamic grasping, which are essential for accurately modeling human and object motion, training practical dynamic grasping strategies, and enabling immediate real-world deployment. To be specific, human hand motions during object handovers are precisely captured using a sophisticated visual system, preserving details of human motions. For the object, precise pose annotations and detailed point cloud representations are provided, effectively capturing the true dynamics of object movement. Additionally, our robotic motion data can be faithfully reproduced in simulators, ensuring true availability.

\noindent\textbf{Human-like Dynamic Grasping Behavior.} We employ a teleoperation system to control a dexterous robotic hand, enabling it to perform grasping actions while seamlessly incorporating human behavioral habits. This approach allows receiver-role robots to learn interaction actions that are more natural and intuitive. During the data collection process, giver-role participants are instructed to hold an object and move it freely in an unconstrained manner, enabling the capture of diverse motion patterns that reflect individual habits. Such data enables robots to better interpret human intentions and adapt their actions accordingly. These advancements are essential for downstream applications involving five-finger dexterous hands and humanoid robots, greatly enhancing robots' human-like intelligence, making robots easier for humans to understand and trust, and paving the way for seamless collaboration.

\section{Method}
In this section, we present our solution \textbf{DynamicGrasp} to the human-to-robot handover problem. Drawing inspiration from how humans receive objects, which contains three key stages: predicting optimal final grasping position and posture; continuously refining approaching poses based on the giver’s movements; and finally grasping object to ensure a stable and successful handover, our solution also consists of 3 stages in ~\cref{fig:pipeline}: (a) \textbf{Grasping Pose Preparation}, (b) \textbf{Approaching Motion Generation}, and (c) \textbf{Goal Pose Alignment}. In the first stage, we train a generative model using both large-scale synthetic data and real-world data to predict potential grasping poses from an observed point cloud, preparing for goal grasp generation in diverse and noisy real-world scenarios. In the second stage, we generate approaching motions by imitating human strategies for interacting with objects. Finally, we adaptively align the dexterous hand pose to the moving object in close proximity to ensure physically plausible and reliable grasping.

Building on this solution and the DexH2R dataset, we establish a benchmark to comprehensively evaluate the effectiveness of methods for the human-to-robot handover task.
In this benchmark, we introduce two technical paradigms for Approaching Motion Generation: the \textbf{auto-regressive approach} and the \textbf{diffusion-policy approach}. Specifically, we implement three baseline methods: MotionNet~\cite{taheri2022goal,liu2024realdex}, Diffusion Policy~\cite{chi2024diffusionpolicy}, and Diffusion Policy 3D~\cite{ze20243d}, adapting them to the dynamic grasping task. Their performance is evaluated on our DexH2R dataset using well-defined metrics that measure motion safety, success rate, trajectory accuracy, and grasp stability. This benchmark provides a robust framework for systematically analyzing the performance of different baseline methods in addressing the human-to-robot handover problem.

\noindent\textbf{Problem Formulation}.
The problem is to predict a sequence of dexterous hand states to grasp a moving object based on historical and current observations of the environment. Let the observation of the state at time step $t$ be defined as $O_t = \{I_t, D_t, P_t^o, S_t^h \}$, where $I_t$ and $D_t$ represent the RGB image and depth image of the scene, $P_t^o$ denotes the point cloud of the object, and hand state $S_t = \{\theta_t^h, q_t\}$. Here, $q_t \in \mathbb{R}^{22}$ represents the articulation pose of the hand and $\theta_t^h$ is the global pose of the hand, which is composed of the global rotation R and global translation T. Let $T_o$ represent the observation time and $T_a$ represent the predicted action time. Given a sequence of observations over the past $T_o=k$ time steps, $\{O_t\}_{t=t_{n-k+1}}^{t_{n}}$, the objective is to generate a sequence of dexterous hand states for the future $T_a=m$ time steps, denoted as $\{S_t\}_{t=t_{n+1}}^{t_{n+m}}$. The generated state $S_t$ can be used to reconstruct the dexterous hand mesh, ensuring safety and stability during grasping. The ultimate objective is to generate reliable human-to-robot handovers in dynamic scenarios.

\subsection{Grasping Pose Preparation}
Given the observed object point cloud, we first generate diverse grasping poses for the dexterous hand using a Conditional Variational Autoencoder (cVAE)~\cite{sohn2015learning,jiang2021hand,kingma2013auto,liu2024realdex}. The cVAE conditions on the object point cloud $P_t^o$ and takes the hand point cloud $P_t^{h}$ as input, encoding them into a latent space using PointNet~\cite{qi2017pointnet}. The decoder then outputs the hand state $S_t$. During inference, we sample the latent space to generate a diverse set of goal-pose candidates.

To ensure grasp stability and safety, we filter the generated poses through a series of checks. First, we perform a stability test in Isaac Gym~\cite{makoviychuk2021isaac}, where forces are applied to the object held by the dexterous hand in six directions. Only poses that maintain the object's stability under these forces are retained. Next, a collision detector is employed to eliminate any dexterous hand poses that intersect with the human hand. Finally, we select the goal state $S_t^{h,g} = \{\theta_t^{h,g}, q_t^{h,g}\}$ that is closest to the current state by minimizing the difference between the global poses $||\theta_t^{h,g} - \theta_t^h||$. Goal pose will be iteratively updated during approaching motion generation and goal pose alignment.

\subsection{Approaching Motion Generation}
\subsubsection{Auto-Regressive Approach}
We take MotionNet~\cite{taheri2022goal,liu2024realdex} as an example of the auto-regressive approach, which was originally developed for motion synthesis of grasping static objects. We extend it to address the dynamic grasping generation task for a dexterous hand by enabling it to predict future trajectories based on historical and current states. 

MotionNet utilizes previous observations of the hand-object interaction scene to auto-regressively predict the future states of the dexterous hand. The input to this module consists of the current hand point cloud $P_t^{h}$, the current velocity of each vertex on the $P_t^{h}$, denoted as $V_t$, the vertex-wise offset from $P_t^{h}$ to the $P_t^{h,g}$, represented by $J_t$ and historical interaction information $\{S_t, P_t^o\}_{t=t_{n-k+1}}^{t_n}$, which includes past hand states and object point clouds.
In an auto-regressive manner, the module predicts the changes in the next hand states $\Delta S_t$ over $T_a$ steps.

Details regarding the training and inference processes can be found in the Supplementary Material.

\subsubsection{Diffusion-Policy Approach}
Diffusion Policy is a method that leverages diffusion models to learn a policy by iteratively denoising trajectory distributions. This approach offers a robust foundation for high-dimensional action synthesis. Notably, unlike MotionNet, Diffusion Policy relies solely on visual observations. The generated goal pose of the dexterous hand is not explicitly considered in this method. Instead, the tendency of object motion is implicitly captured within the visual observations.

Based on the type of observation, two variations of the Diffusion Policy approach were proposed: 
\textbf{(1) Diffusion Policy}~\cite{chi2024diffusionpolicy}: This method operates on 2D RGB images, leveraging visual features to infer object states through projective geometry. However, it faces inherent limitations in handling occlusions and depth ambiguity due to the lack of explicit spatial information.  
\textbf{(2) Diffusion Policy 3D}~\cite{ze20243d}: This method overcomes these constraints by using 3D point clouds as input observations. The explicit spatial encoding of point clouds allows for direct reasoning about object geometry and contact dynamics, which is crucial for tasks that require precise dexterous grasp synthesis.

To adapt Diffusion Policy for our tasks, we enhance its input by incorporating depth images, allowing the model to capture 3D structural information. Consequently, the observation for Diffusion Policy is defined as $\{I_t, D_t, S_t\}$, whereas for Diffusion Policy 3D, it is defined as $\{P_t^o, S_t\}$. At each time step, the observations from the previous $k$ frames are used as the conditioning input for the diffusion process~\cite{ho2020denoising}, which iteratively denoises the trajectory over the future $m$ frames.
We adopt the training strategy proposed in the original Diffusion Policy~\cite{chi2024diffusionpolicy} and its 3D counterpart~\cite{ze20243d}.

\subsection{Goal Pose Alignment}

In this stage, we iteratively update the position of the dexterous hand to bring it closer to the final stable grasping state.
Specifically, this approach utilizes the {Grasping Pose Preparation} module to generate validated poses based on the current object point cloud. We then align the future dexterous hand pose to closely match the goal pose. The alignment process is structured as follows:
First, we calculate the global translation difference, denoted as $d$, between the current hand point cloud $P_t^h$ and the goal hand point cloud $P_t^{h,g}$. If $d < U$, we begin doing goal pose alignment, $U$ represents the interpolation threshold. Assuming motion is linear in a short time period, we estimate the number of frames $N$ needed to smoothly reach the goal pose as $N = d/v$, where $v$ is the predefined velocity (set to $2$cm per frame in practice).
Next, we linearly interpolate between the current hand pose and the goal pose.
Finally, after each step, we iteratively update the state of the hand and object and recompute the distance $d$ until it falls below a threshold of $1$cm, indicating that the hand has successfully aligned with the reliable grasping goal. A collision check is incorporated throughout the process to ensure safe and effective alignment.
Through goal pose alignment, the proposed method achieves reliable, collision-free, and dynamic grasping. It is particularly effective in close proximity scenarios.

\section{Experiments}
The experiments were designed to assess the performance of our proposed solution to the human-to-robot handover problem, focusing on key metrics such as safety, precision, and reliability. To facilitate comprehensive evaluation, we introduce a benchmark that compares the performance of three baseline methods in dynamic dexterous grasping using the DexH2R dataset, highlighting their respective strengths and limitations. The evaluations were conducted in both simulated and real-world environments to rigorously validate the robustness and practicality of the proposed system.
\subsection{Experiment Setup}
We train MotionNet, Diffusion Policy, and Diffusion Policy 3D on an A40 GPU server. We use Isaac Gym as the simulation environment. The dataset is split into a training set (2,888 trajectories with 46 objects and 32 subjects), a validation set (591 trajectories), and a test set (803 trajectories), which comprises three types of instances: unseen objects/unseen subjects, unseen objects/seen subjects, and seen objects/unseen subjects. Complete partitioning details are provided in the appendix. The two diffusion-based models are trained for 50 epochs, whereas MotionNet requires 500 epochs to converge. MotionNet required 14 GPU days of training, while Diffusion Policy and Diffusion Policy 3D each took approximately 5 GPU days.

\subsection{Evaluation Metrics}
To evaluate the quality, robustness, and safety of grasping poses and approaching methods, we define the following metrics in ~\cref{tab:evaluation_metrics}, with detailed calculations in the supplementary material. These provide a comprehensive benchmark for quantitative comparisons and highlight areas for improvement. The table below summarizes the metrics, with detailed calculations in the supplementary material.

\begin{table}[ht]
\centering
\resizebox{\columnwidth}{!}{ 
\begin{tabular}{p{2cm}|l|p{6cm}} 
\hline
\textbf{Category} & \textbf{Metric} & \textbf{Description} \\ \hline
\multirow{4}{*}{\centering Grasping Pose} & \textbf{Success Rate (succ1)} & Success rate under a single random force applied to the object. \\ \cline{2-3}
 & \textbf{Success Rate6 (succ6)} & Success rate under random forces in six directions. \\ \cline{2-3}
 & \textbf{Penetration Depth (pen\_dep)} & Measures hand penetration into the object surface, penalizing deep penetration {The unit is in centimeters (cm)}. \\ \cline{2-3}
 & \textbf{Diversity (div)} & Standard deviation of pose parameters across successful grasps. \\ \hline
\multirow{6}{*}{\centering Approaching} & \textbf{Success Rate (succ)} & Percentage of trajectories achieving the target grasping pose. \\ \cline{2-3}
 & \textbf{Trajectory Length (traj\_len)} & Total length of the generated trajectory (ideal range: 1--2 meters). \\ \cline{2-3}
 & \textbf{Total Infer Frames (infer\_fr)} & Average frames required to infer a trajectory (lower is better). \\ \cline{2-3}
 & \textbf{Penetration Depth (pen\_dep)} & Depth of hand penetration into the object (smaller is safer), { The unit is in centimeters (cm)}. \\ \cline{2-3}
 & \textbf{Penetration Frame (pen\_fr)} & Frames where penetration exceeds a safety threshold. \\ \cline{2-3}
 & \textbf{Safety Rate (safe)} & Percentage of trajectories with mean penetration below the threshold. \\ \hline
\end{tabular}
}
\caption{Summary of Evaluation Metrics.}
\label{tab:evaluation_metrics}
\end{table}

\subsection{Evaluation for Grasping Pose Generation}
\renewcommand{\arraystretch}{0.85}
\begin{table*}[t]
\centering
\resizebox{\linewidth}{!}{ 
    \begin{tabular}{@{}c|ccccc|ccccc@{}}
    \toprule
    \multirow{2}{*}{Method} & \multicolumn{5}{c|}{Easy Mode (U=10cm)} & \multicolumn{5}{c}{Hard Mode (U=5cm)} \\ 
    \cmidrule(l){2-6} \cmidrule(l){7-11}
                         & \textbf{succ $\uparrow$} & \textbf{safe $\uparrow$} & \textbf{pen\_dep $\downarrow$} & \textbf{pen\_fr $\downarrow$} & \textbf{infer\_fr $\downarrow$} 
                         & \textbf{succ $\uparrow$} & \textbf{safe $\uparrow$} & \textbf{pen\_dep $\downarrow$} & \textbf{pen\_fr $\downarrow$} & \textbf{infer\_fr $\downarrow$} \\ \midrule
    \multirow{1}{*}{MotionNet}    & 71.1 & 25.9 & 0.72 & 8.3 & 87.4 & 26.6 & 15.4 & 0.79  & 14.5  & 99.6 \\
    \multirow{1}{*}{DP}     &39.4   &42.6 & 0.69 & 6.8  & 97.7  & 8.3   & 38.5  & 0.78  & 9.7  & 101.2 \\
    \multirow{1}{*}{DP3}  & 66.3  & 50.1 & 0.73 & 5.0 & 96.8 & 27.1 & 33.7 & 0.84 & 9.3 & 103.7 \\ \bottomrule
    \end{tabular}
}
    \caption{Comparison for Dynamic grasping solutions on test dataset.}
    \label{tab:seen_obj}
\end{table*}
We evaluate grasping pose generation on the \textbf{DexH2R} dataset, as shown in ~\cref{tab:grasp_generation}, which includes 56 objects and 79,911 grasping poses. Our model is first pre-trained on the synthetic \textbf{DexGraspNet} dataset (5,355 objects, 1.32M poses) and then fine-tuned on \textbf{DexH2R}. We compare two baselines, \textbf{cVAE}~\cite{liu2024realdex} and \textbf{DexgraspAnything}~\cite{graspanyting}, analyzing grasp success rate, diversity, and penetration depth. Results show that training directly on \textbf{DexH2R} achieves comparable performance to pretraining on \textbf{dexgraspnet}~\cite{wang2023dexgraspnet}, with higher \textbf{succ1} (89.38 vs. 88.75) and \textbf{succ6} (35.56 vs. 35.00). This highlights the effectiveness of \textbf{DexH2R}'s grasping poses, despite its focus on dynamic human-robot handover tasks. An ablation study is conducted to assess the impact of pretraining on synthetic data. All experiments are performed on an NVIDIA A40 GPU using the Adam optimizer.

\begin{table}[h]
\centering
\Large 
\resizebox{\columnwidth}{!}{
\begin{tabular}{c|c|cccc} 
\toprule
\textbf{Method} & \textbf{w/ pretrain} & \textbf{succ6 $\uparrow$} & \textbf{succ1 $\uparrow$} & \textbf{div $\uparrow$} & \textbf{pen\_depth $\downarrow$} \\ 
\midrule
\multirow{2}{*}{\centering \textbf{cVAE}~\cite{liu2024realdex}} 
 & \cmark & 35.00 & \textbf{79.38} & \textbf{0.134} & \textbf{2.23}  \\
 & \xmark & \textbf{35.56} & 76.56 & 0.105 & 2.36  \\
\midrule
\multirow{2}{*}{\centering \textbf{DexgraspAnything~\cite{graspanyting}}} 
  & \cmark & \textbf{52.81} & 88.75 & \textbf{0.123} & \textbf{1.64}\\
 & \xmark & 50.62 & \textbf{89.38} & 0.112 & 1.83  \\
\bottomrule
\end{tabular}}
\caption{Evaluation results on the \textbf{DexH2R} dataset. We compare cVAE with DexgraspAnything and conduct an ablation study on pretraining with large-scale synthetic data.}
\vspace{-7pt}
\label{tab:grasp_generation}
\end{table}

\begin{figure}[h]
    \centering
    \includegraphics[width=\linewidth]{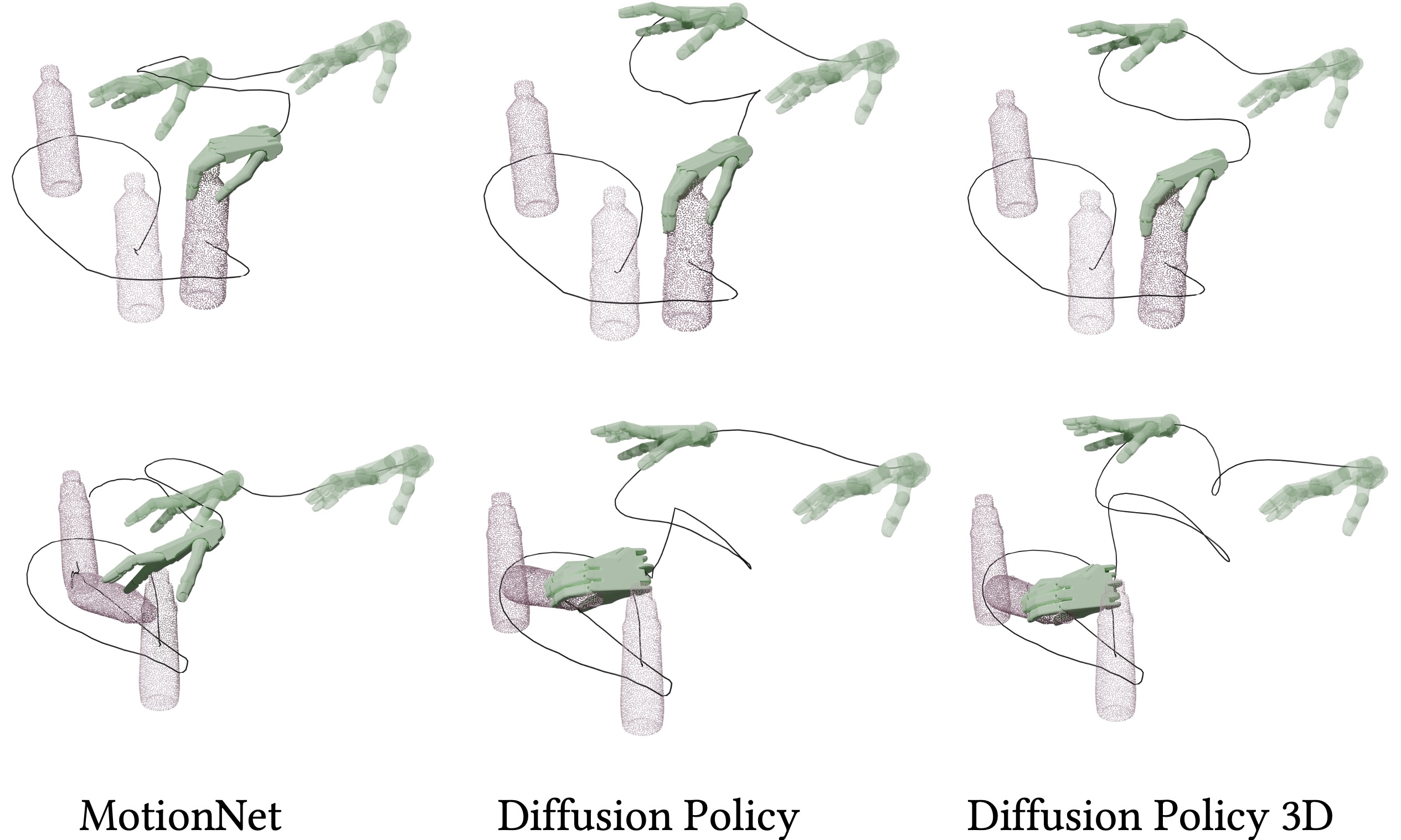}
    \caption{The qualitative comparison of three baseline methods on approaching motion generation.}
    \vspace{-1em}
    \label{fig:result}
\end{figure}

\subsection{Evaluation for Approaching Motion Generation}  
We evaluated three approaching motion generation algorithms (MotionNet, DP, and DP3) on the test set combining unseen subjects and objects, as summarized in ~\cref{tab:seen_obj}. We also include a qualitative comparison of three baseline methods, as shown in~\cref{fig:result}. To ensure safe, reliable, and applicable motion synthesis in real-world scenarios, we used \textbf{Goal Pose Alignment} as a corrective strategy.

To comprehensively evaluate the models' robustness and precision, we defined two task modes: \textbf{Easy Mode} (\( d \leq U = 10 \) cm) and \textbf{Hard Mode} (\( d \leq U = 5 \) cm). Here, \( d \) is the global translation difference computed between the current pose of the dexterous hand (\( P_1 \)) and the goal grasping pose (\( P_2 \)). The thresholds $U$ of 5 cm and 10 cm represent the initial distances at which the goal pose alignment process begins. The Easy Mode evaluates the model's ability to understand the global position of the object, while the Hard Mode tests fine-grained pose alignment and collision avoidance in close proximity, crucial for safe and precise interactions. By incorporating both modes, our benchmark provides a balanced evaluation of model robustness, ensuring applicability in real-world scenarios where both global localization and local precision are essential.

\subsubsection{Model Performance Comparison and Analysis}
Here we present the performance analysis on three baseline methods in easy mode and hard mode settings.
\begin{itemize}
    \item \textbf{Easy Mode}:
        \textbf{MotionNet} outperformed Diffusion Policy (DP) and Diffusion Policy 3D (DP3) in terms of success rate. This is likely because MotionNet adopts a more aggressive strategy, prioritizing quick object grasping. However, this approach often neglects collision avoidance, resulting in lower safety rates and higher penetration frames.
        \textbf{DP3} balances grasping with collision avoidance, achieving a better safety rate compared to MotionNet.
        \textbf{DP} adopts a conservative strategy, avoiding collisions by approaching the object slowly. While this improves safety, it often results in failure to grasp within the specified frame limit.

    \item \textbf{Hard Mode}:
        \textbf{DP3} achieved a success rate comparable to \textbf{MotionNet} and outperformed \textbf{DP}, while maintaining a high safety rate. This indicates that \textbf{DP3} can better perceive the object's precise position through its point cloud input and encoder, ensuring both successful grasping and safety.
\end{itemize}

\section{Conclusion}
In this paper, we introduce DexH2R, a comprehensive real-world dataset for human-to-robot handovers, developed using the ShadowHand dexterous robotic hand. Our dataset encompasses a diverse range of interactive objects and human participants, rich multi-modal sensor data, detailed annotations, and human-like robotic motions. Additionally, we propose an effective solution for generating human-like dynamic grasping motions in human-to-robot handover scenarios, accompanied by a thorough evaluation and analysis. We believe our benchmark will significantly advance research in human-to-robot handovers.

\section{Acknowledgements}
This work was supported by NSFC (No.62206173), Shanghai Frontiers Science Center of Human-centered Artificial Intelligence (ShangHAI), and MoE Key Laboratory of Intelligent Perception and Human-Machine Collaboration (KLIP-HuMaCo).

{
    \small
    \bibliographystyle{ieeenat_fullname}
    \bibliography{main}
}

\end{document}